\documentclass{article}

\usepackage{PRIMEarxiv}

\usepackage[utf8]{inputenc} 
\usepackage[T1]{fontenc}    
\usepackage{hyperref}       
\usepackage{url}            
\usepackage{booktabs}       
\usepackage{amsfonts}       
\usepackage{nicefrac}       
\usepackage{microtype}      
\usepackage{lipsum}
\usepackage{fancyhdr}       
\usepackage{graphicx}       
\graphicspath{}     
\usepackage{algorithm,algpseudocode}
\usepackage{algpseudocode}
\usepackage{fontawesome}
\usepackage{graphicx}
\usepackage{amsmath}
\usepackage{multirow}

\title{Hybrid CNN -Interpreter: Interprete local and global contexts for CNN-based Models
}

\author{
  Wenli Yang\\
  School of Information and Communication Technology \\
  University of Tasmania \\
  Australia\\
  \texttt{\{wenli.yang\}@utas.edu.au} \\
   \And
  Guan Huang \\
  School of Information and Communication Technology \\
  University of Tasmania \\
  Australia\\
  \texttt{\{guan.huang\}@utas.edu.au} \\
  \AND
  Renjie Li \\
  School of Information and Communication Technology \\
  University of Tasmania \\
  Australia \\
  \texttt{\{renjie.li\}@utas.edu.au} \\
   \And
   Jiahao Yu \\
  School of Information and Communication Technology \\
  University of Tasmania \\
  Australia \\
  \texttt{\{jiahao.yu\}@utas.edu.au} \\
   \And
  Yanyu Chen \\
  School of Information and Communication Technology \\
  University of Tasmania \\
  Australia \\
  \texttt{\{yanyu.chen\}@utas.edu.au} \\
  \And
  Quan Bai \\
  School of Information and Communication Technology \\
  University of Tasmania \\
  Australia \\
  \texttt{\{quan.bai\}@utas.edu.au} \\
   \And
  Byeong Kang \\
  School of Information and Communication Technology \\
  University of Tasmania \\
  Australia \\
  \texttt{\{byeong.kang\}@utas.edu.au} \\ 
}

\begin{document}
\maketitle

\begin{abstract}
Convolutional neural network (CNN) models have seen advanced improvements in performance in various domains, but lack of interpretability is a major barrier to assurance and regulation during operation for acceptance and deployment of AI-assisted applications. There have been many works on input interpretability focusing on analyzing the input-output relations, but the internal logic of models has not been clarified in the current mainstream interpretability methods. In this study, we propose a novel hybrid CNN-interpreter through: (1) An original forward propagation mechanism to examine the layer-specific prediction results for local interpretability. (2) A new global interpretability that indicates the feature correlation and filter importance effects. By combining the local and global interpretabilities, hybrid CNN-interpreter enables us to have a solid understanding and monitoring of model context during the whole learning process with detailed and consistent representations. Finally, the proposed interpretabilities have been demonstrated to adapt to various CNN-based model structures.
\end{abstract}


\maketitle

\begin{figure*}[ht]
\centering
\includegraphics[width=1.1\textwidth]{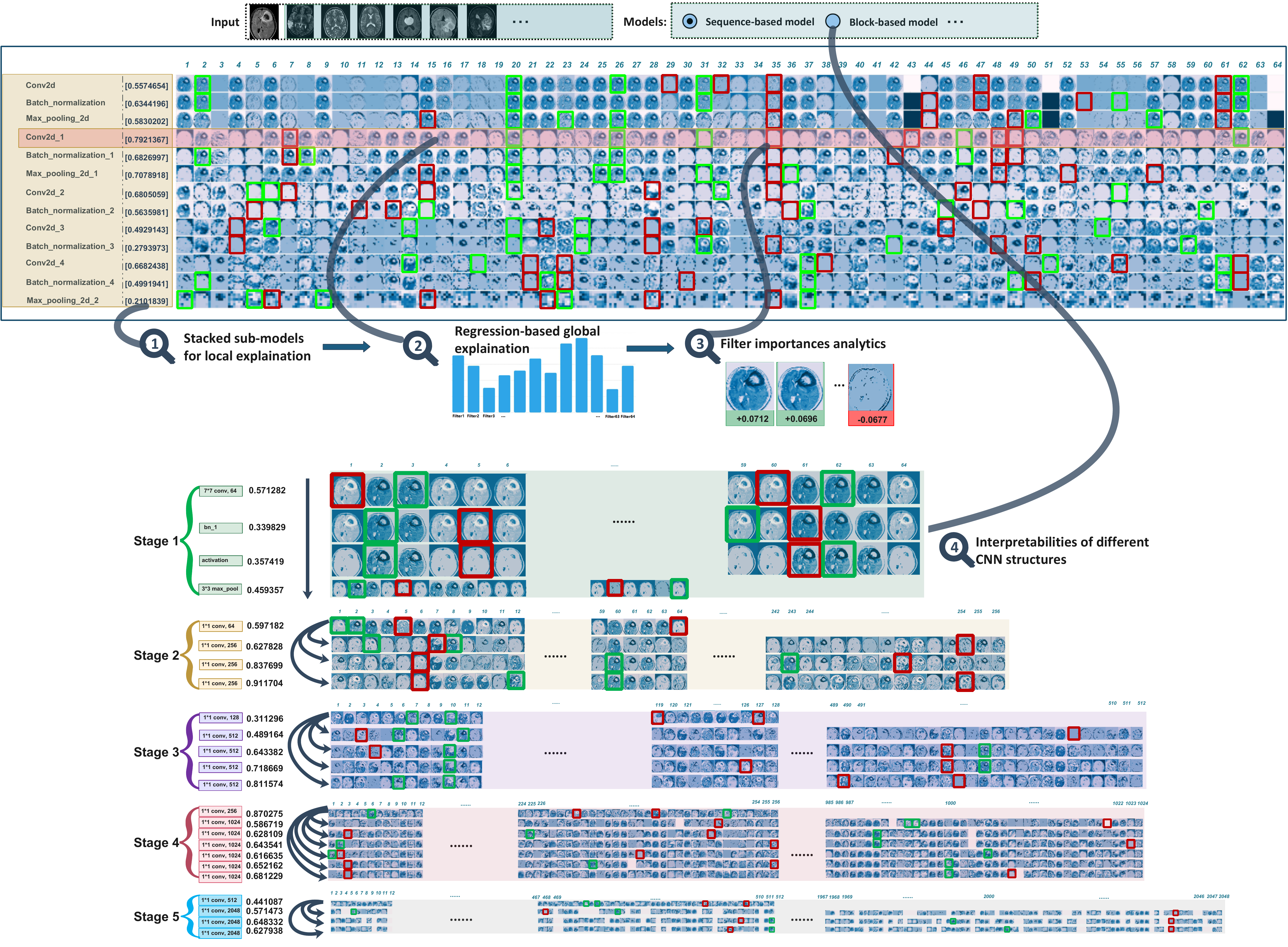}
\caption{\textbf{Hybrid CNN-interpreter: understand how different convolutional neural networks learn through layers and filters:} \textbf{1)} Local interpretability to represent each layer's output and learning ability; \textbf{2)} Global interpretability to explore the representations ability of feature maps(learned by different filters) by using regression models; \textbf{3)} feature importance analytic to indicate the contributions of convolution filters when making predictions. Overall, it enables for both local and global interpretability for different CNN structures.} 
\label{framework1}
\end{figure*}

\section{Introduction}

Although the performance of convolutional neural network (CNN) models has significantly increased over the past decade, yet without reliable interpretabilities that effectively represent the learning processes, humans still consider untrust¬worthy when continue to view CNN models to be used as unreliable when utilized for real-world decision making \cite{liu2021trustworthy}. The lack of trust undermines the deployment of CNN-based technologies into many domains, such as medical diagnosis \cite{zhang2022applications}, healthcare \cite{pawar2020explainable}, autonomous driving \cite{atakishiyev2021explainable}, etc. This motivates the need for building trust in these CNN-assisted decision making. Building trust by adding interpretabilities has significant theoretical and practical value for both improving model performance and enhancing transparency. Interpretabilities can be represented by using a variety of expressions, such as visual interpretability and semantic interpretability  \cite{zhang2018visual}. Among different expressions, the visual interpretability of CNN-based models is the most fundamental and direct way to explain the network representations.

In terms of visual interpretability methods, the interpretability of CNN-based models emphasizes either visualization of training data rules \cite{binkowski2018demystifying} or the visualization inside the model \cite{zhang2018visual}. Presently, most visual interpretability methods focused more on the understanding of input features on the model performance ensured rather than on the context of CNN-based models, which can reveal the learning process through each layer and the generic features learned from any black box models. Additionally, the internal logic of models has not been clarified in the current mainstream visual interpretability methods. Most existing internal logical discussion about models is used for tree-based models \cite{waghen2021multi}. For CNN-based models, to the best of our knowledge, correlation interpretabilities have not been discussed so far.

We present a novel hybrid CNN interpreter, which builds stacking ensemble models of each layer for local interpretabilities and extends local interpretabilities to compute global correlation to assess the importance of convolution filters. The results are illustrated by binary classification of brain tumor data in the different model structures. It makes three innovative improvements:
\begin{enumerate}
\item {Build an original stacking forward propagation algorithm by computing the contributions of each layer to the final prediction. In stacking, each time takes the outputs of the specific layer as input and connects to the final probability mapping layer directly, which can use the set of predictions as a local context and examine feature maps learned by different layers' ability for the final prediction contributions.}
\item {Extend local interpretabilities to a global interpretation by layer-based regression models, which are constructed by using all the local interpretabilities as dependent variables and each feature map representation as an independent variable across the entire dataset. This enables the examination of the representation ability of feature maps (learned by different filters) for interpreting a model’s global behavior.} 
\item{Extract filter importance by assigning scores to each filter in each layer of a model that indicates the correlation of convolution filters when making a prediction, which can help people understand the relationship between the filters and the target variable and can conditionally improve the performance of the model, potentially making the model lighter and speeding up the model’s working by removing the unimportant filters.}
\item {Applicable for a broad range of CNN-based models, which can be utilized to interpret both sequence and nonsequence-focused models. For the sequence models, the interpretabilities can be through each layer to provide the learning context, whereas for the nonsequence models, the interpretabilities can be customized for different stages, blocks, or specific layers, etc. This reveals how different learning is in different models and in different training processes.}
\end{enumerate}

\section{Related Work}
\subsection{Convolutional neural network structures}

CNNs with strong representation ability of deep structures have ever-increasing popularity in many applications. Figure 1 shows the historical evolution of various CNN models. AlexNet \cite{krizhevsky2017imagenet} is a leading structure of convolutional neural networks and has huge applications for classification tasks. The evolution after the AlexNet can be mainly summarized in two ways: sequence-focused models by increasing the depth of networks and nonsequence-focused models by adding units or modules. 

\begin{figure*}[ht]
\centering
\includegraphics[width=1.0\textwidth]{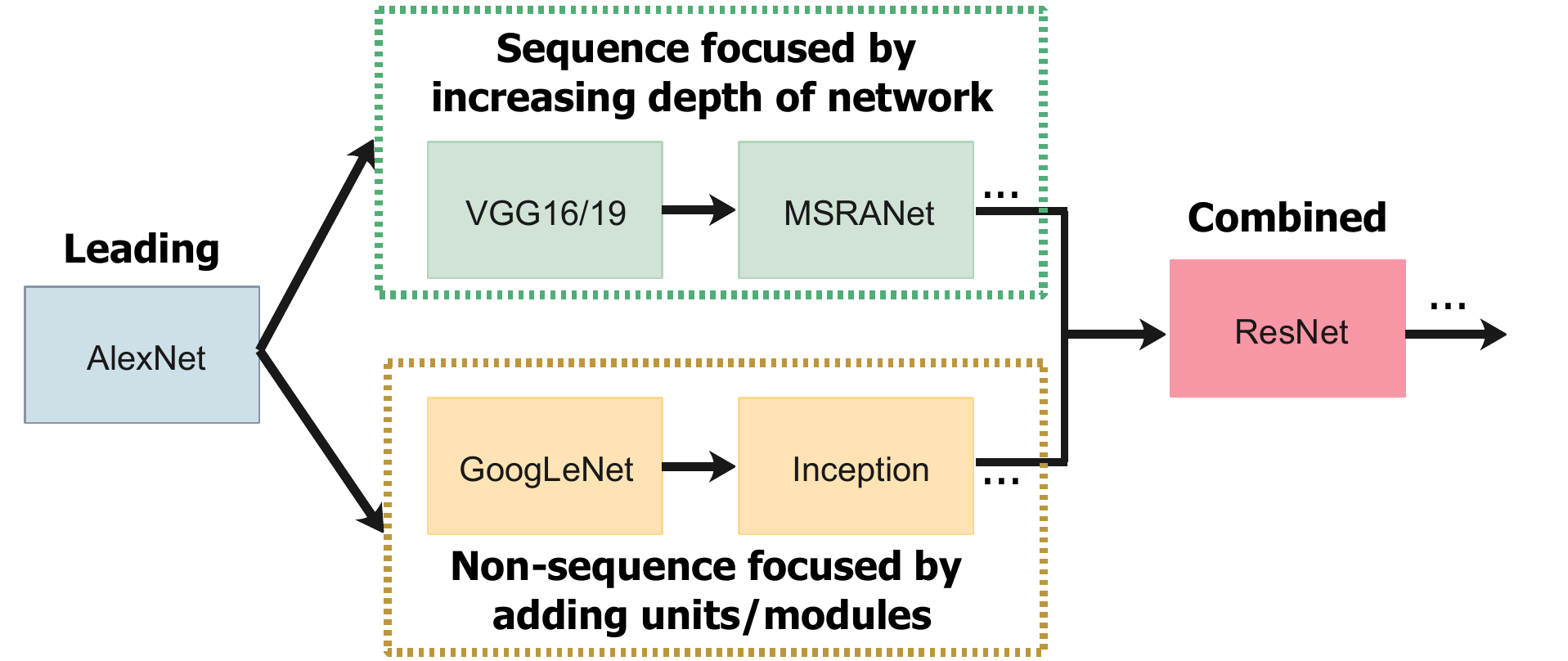}
\caption{\textbf{Summarization of CNN structures}}
\label{resnet_regression}
\end{figure*}

\textbf{Sequence focused models}: by leveraging the sequence information through layers to improve the model performance, such as VGG16 and VGG19  \cite{simonyan2014very}, MSRANet \cite{zheng4178119msranet}. All of these models are focused on using multiple convolutional layers and activation layers to increase the depth of network, which can extract deeper and better features than the simple structure.

\textbf{Non-sequence-focused models}: by using modular structures to add functional units or components to extend the width of network. Typical examples such as GoogLeNet \cite{szegedy2015going}, Inception \cite{szegedy2016rethinking}. GoogLeNet utilizes multiple branches, which allows the network to choose between multiple convolutional filter sizes in each branch. An inception network comprises repeating components referred to as inception modules to increase the representational power of a neural network.

By combining them together, ResNet \cite{he2016deep} uses four stages made up of residual blocks, each of which uses several residual blocks with convolutional blocks and identity blocks. Finally, residual blocks are stacked together on top of each other to form the whole network, which can gain better performance from considerably increased depth.

In this paper, we will pick up AlexNet and ResNet as two representative models to interpret the learning process of different model structures.

\subsection{Visual interpretabilities of CNN-based models}

According to different implementation methods, the visual interpretability methods can be divided into three different categories: input visualization method, model visualization method, and mixed visualization method. Table 1 states the overall summary of possible explainable methods.

\begin{table}[]
\caption{Different types of visual interpretability methods}
\label{tab:my-table}
\begin{tabular}{|l|l|}
\hline
\begin{tabular}[c]{@{}l@{}}Types of Interpretable\\    \\ Method\end{tabular} & Description                                             \\ \hline
Input   Visualization &
  \begin{tabular}[c]{@{}l@{}}Provides an accessible way to view and understand the impact \\ of initial input data on the final model performance.\end{tabular} \\ \hline
Model Visualization                                                         & Provides analytics of layer-based outputs inside models \\ \hline
Mixed Visualization &
  \begin{tabular}[c]{@{}l@{}}Provides both input visualization and model visualization, \\ which focused on building connections or correlations.\end{tabular} \\ \hline
\end{tabular}
\end{table}

The input visualization explains processes simultaneously from the initial input stage to the final output result. For example, Jeyaraj and Nadar \cite{jeyaraj2019computer} stated a regression-based partitioned method for oral cancer diagnosis. The network was demonstrated with two partitioned layers for labelling and categorized a multidimensional hyperspectral picture by tagging the region of interest. It handled feature maps with little variation and complex vector feature maps. Another example was the importance estimation network produced by Gu et al. \cite{gu2020vinet}, which was diagnosed by the classification network by investigating the irrelevant information. The model aims to detect the most significant sections of the original input images and provided an accurate diagnosis after being trained with the proper regularization settings. The input visualization merely provides a user-friendly interface for seeing and understanding input data, but no information on how the relevant features contribute to the prediction is supplied.

Generally, model visualization is based on the level of layers and finds out the prediction results of the model between different layers. For example, Graziani et al.\cite{graziani2020concept} employed “The concept activation vector (TCAV)” to transform medical pictures into quantitative characteristics by using radiomics, which focused on explaining the predicted output globally according to high-level visual features. Additionally, Villain et al. stated a novel GradCAM method on brain MRI image datasets \cite{villain2021visual}. To identify the regions of interest by the visual convolutional neural network models, the authors claimed that the output of every convolution layer was collected by the global average pool layer and merged to obtain a single activation map to visualize. The model visualization could find hidden problems inside the model, but the currently selected features used for model interpretability may not be repeatable in each input. 

Hybrid visualization model refers to a combination of both input visualization and model visualization. This model tries to focus on not only explaining the local prediction but also exploring global knowledge inside the model. Hybrid visualization model concentrates on finding the connections or correlations between the input aspect and model layer aspect. Unfortunately, to my knowledge, there are no obvious research on mixed visualization method.

In this paper, we discuss how to interpret CNN-based models using hybrid visualization to make the interpretability appropriate for different model structures. We describe the local interpretability by building a set of stacked ensemble models to provide various input visualization. Then, we extend local interpretability to model visualization by using regression-based analytics. Finally, we identify the importance of each filter in each layer to represent the global context of the models. 

\section{Method}
Hybrid CNN-interpreter aims to explain the deep learning model's different layers and the filter's feature representation ability for the final prediction contribution. It provides global insight into not only the model's layer level, but also the filter level along that layer. The output of the hybrid CNN-interpreter is a layer and filter-based importance distribution matrix, representing the importance level of feature maps learned from different layers and filters by the model. The hybrid CNN-interpreter consists of the original CNN-model forward propagation module, linear regression module and filter importance analysis module.

\begin{figure*}[htbp]
\centering
\includegraphics[width=0.98\textwidth]{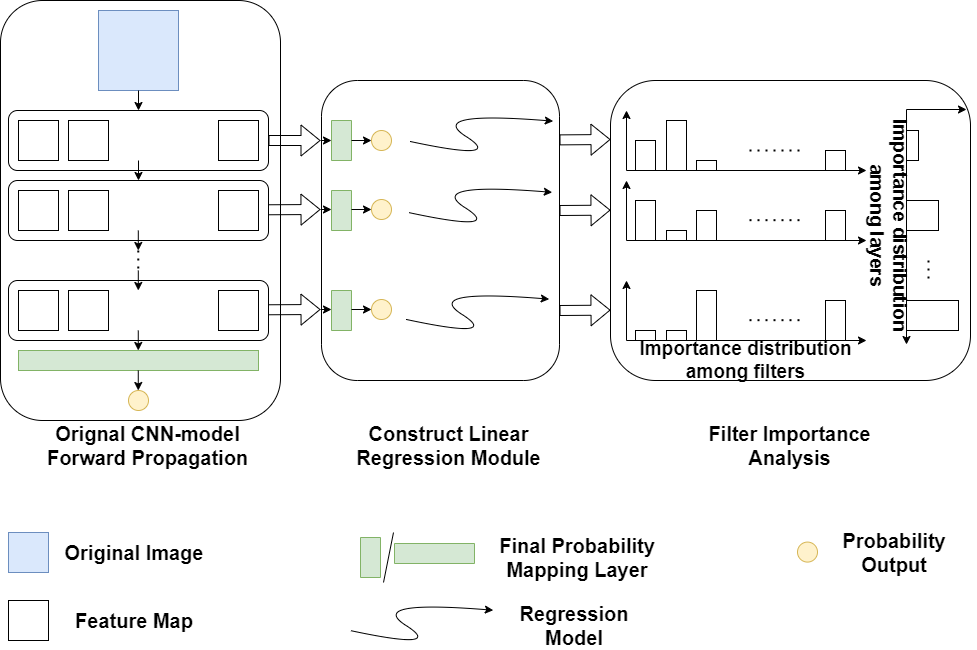}
\caption{\textbf{Hybrid CNN-interpreter framework}: including Original CNN-model Forward Propagation, Linear Regression Module and Filter Importance Analysis.}
\label{fig:whole_structure}
\end{figure*}

\subsection{Stacking Forward Propagation}
In the first original forward propagation module, each image is sent to the original model. We skip and connect each layer ($l_{i}(X_{ij})$) directly to the final probability mapping layer ($f(\cdot)$), and the output probability ($Y_{i}$) is recorded. This is to examine feature maps learned by different layers' ability for the final classification contribution (Equation~\ref{eq:forward}).
\begin{equation}
\begin{aligned}
    &f[l_{i}(X_{ij})]= Y_{i} \label{eq:forward}
\end{aligned}
\end{equation}
where $X_{ij}$ represents the $j^{th}$ feature map at the $i^{th}$ layer.

\subsection{Linear Regression Module}
In the second step, a linear regression model is constructed by using the output probability as a dependent variable and each feature map's mean value as independent variables. This is to examine feature maps' (learned by different filters) representation ability for the classification (Equation~\ref{eq:regression}). 
\begin{equation}
\begin{aligned}
    &Y_{i}=b_{i0}+\sum{b_{ij}\overline{X}_{ij}} \label{eq:regression}
\end{aligned}
\end{equation}
where $\overline{X}_{ij}$ represents the mean value of feature map $X_{ij}$.
We applied L2 regularization and the objective function is represented in Equation~\ref{eq:l2}.
\begin{equation}
\begin{aligned}
    &L=\sum{(\hat{Y_{i}}-Y_{i})^{2}+\lambda\sum{b_{ij}^{2}}} \label{eq:l2}
\end{aligned}
\end{equation}
where $\hat{Y_{i}}$ is the predicted value and $Y_{i}$ is the true value.

Algorithm \ref{alg:export-feature} indicates the process of how $\overline{X}_{ij}$ are extracted from the model outputs. 

\begin{algorithm}
  \caption{Mean values of feature maps extraction}
  \label{alg:export-feature}
  \begin{algorithmic}[1]
    \Procedure{export-features}{$img,lnames$}\Comment{ The input of this algorithm are images and the layer names of the model}
        \State $model.predict(img) \gets features$ \Comment{ The feature maps are obtained by visualisation the model prediction.}
        \For{$lname, features$} 
            \State $mean$-$list =$ [ ]
                \For{$i$ in range($features$-$number$) }
                    \State $x$ = $features$[0, :, :, i]
                    \State $x$ = $x$ - $x$.mean() \Comment{$x$.mean() the mathematical mean value of $x$}
                    \State $x$ = $x$ / $x$.std() \Comment{$x$.std() the mathematical standard deviation value of $x$}
                    \State $x$ = $x$ * 32
                    \State $x$ = $x$ + 64
                    \State $x$ = numpy.clip($x$, 0, 255) \Comment{Use numpy.clip() function to limit values outside the interval are clipped to the interval edges}
                    \State $mean$-$list$.append($x$) \Comment{ Store the mean value of the feature maps into $mean$-$list$}
                \EndFor
        \EndFor
      \State \textbf{return} $mean$-$list$\Comment{The output of the algorithm is a list with the mean value of the feature maps}
    \EndProcedure
  \end{algorithmic}
\end{algorithm}

In our experiment, we use Ridge regression model \cite{hoerl1970ridge} to calculate the variance of our feature maps. The Ridge model solves regression problems with an $L2$ regularization loss function, the equation of the loss is shown in Equation \ref{eq:l2}. In the Ridge model, the parameter $\alpha$ ($\alpha$ = 1 in our experiment) is used to control the regularization strength. The input of our regression model is the mathematical mean value of our feature maps and the output of the CNN models after prediction (confidence score in our experiment). Algorithm~\ref{alg:regression-algorithm} shows the details of 
\begin{algorithm}
  \caption{Regression Algorithm}
  \label{alg:regression-algorithm}
  \begin{algorithmic}[2]
    \Procedure{regression}{$layer$-$names$,$f$-$mean$,CNN-predictions}\Comment{ The input of this algorithm are mean values of the feature maps and the layer names of the model}
        \State x = $f$-$mean$
        \State y = CNN-predictions
        \State model = Ridge(alpha=1.0) \Comment{ The model is loaded from sklearn library}
        \State predictions = model.predict(x)
        \State coefficients, standard-errors, t-values, p-values = sklearn-math-function(y, predictions,features) \Comment{we use a combination of mathematical tools from sklearn library to calculate coefficients, standard-errors, t-values and p-values}
      \State \textbf{return} coefficients, standard-errors, t-values, p-values
    \EndProcedure
  \end{algorithmic}
\end{algorithm}

Then, the regression model will return four parameters to represent the importance of our feature map, i.e. the coefficient of determination (${R}^{2}$), standard-errors (SEs),  t-values and p-values. The equation of ${R}^{2}$ is demonstrated in \ref{eq:r_sqaured}. The meaning of ${R}^{2}$ is the proportion of the variance of the variable that is explained by this predicted model. In our experiment, we use ${R}^{2}$  to evaluate the reliability of the prediction, if the value of ${R}^{2}$ close to 1, which means the model predictions are reliable and effective. Conversely, if the value of ${R}^{2}$ is close to 0, the predictions of the model are not reliable and ineffective. 
\begin{equation}
\begin{aligned}
    R^{2}=\dfrac{\sum _{i}e_{i}^{2}}{\sum _{i}\left( y_{i}-\overline{y}\right) ^{2}} \label{eq:r_sqaured}
\end{aligned}
\end{equation}
The equation of SE is shown in Equation \ref{eq:SE}. The SE is another statistical tool for us to evaluate the degree of deviation of the sample mean from the overall true mean. Normally, a higher SE value indicates that the mean value of the feature maps is more distinct, whereas a smaller SE value means the features extracted are similar.
\begin{equation}
\begin{aligned}
SE=\dfrac{\sigma }{\sqrt{n}}
\label{eq:SE}
\end{aligned}
\end{equation}
where $SE$ is the standard error of the sample, $\sigma$ is the sample standard deviation and $n$ is the number of samples.

\subsection{Filter Importance Analysis Module}
In the third filter importance analysis module, a matrix of different filters' representation scores has been calculated, the matrix helps people understand the contribution of the feature map learned by different filters in each layer of the network to the final classification result.

In our settings, the t-value (Equation \ref{eq:t-value}) is used to indicate the level of difference between features. The t-value is calculated as the difference expressed in standard error units. It means that the standard deviation estimated is far away from 0. A large t-value indicates the evidence against the null hypothesis, in other words, we could declare a relationship between the CNN-predictions and the feature maps.
\begin{equation}
\begin{aligned}
t=\dfrac{\overline{x}-\mu _{0}}{\dfrac{s}{\sqrt{n}}}
\label{eq:t-value}
\end{aligned}
\end{equation}
where $\overline{x}$ is the sample mean, $\mu _{0}$ is the population mean, $s$ is the sample standard deviation and $n$ is sample size.

We also used p-value (Equation \ref{eq:p-value}) to evaluate two variables, namely, they are the importance of the features and the CNN model prediction accuracy. We want to examine the relationship between the predictor variables and the response variable to find out if higher accuracy in model prediction will result in higher importance values in the feature maps' mean values.
\begin{equation}
\begin{aligned}
p=\dfrac{\widehat{p}-p0}{\dfrac{\sqrt{p_{0}\left( 1-p_{0}\right) }}{n}}
\label{eq:p-value}
\end{aligned}
\end{equation}
where $\widehat{p}$ is sample proportion, $p_{0}$ is assumed population proportion in the null hypothesis and $n$ is the sample size.

\section{Experiment and Discussion}
\subsection{Data preparation}
A total of 253 public brain MRI images dataset (155 tumorous images and 98 nontumorous images) are used to build the classification models. Since CNNs can be independent of translation, view position, size, and lighting \cite{yang2020visual}, the data augmentation was applied by manually flipping and rotating the image sets. Moreover, considering the dimension of the image may be changed after rotation \cite{hu2020efficient}, the image will be only rotated arbitrarily between 0 and 10 degrees in our experiments. 

Finally, after the above data augmentation, the dataset has 1,085 positive and 980 negative examples. We use 70\% of the images for training and 15\% of the images for validating to generate the AlexNet and ResNet classifier models. The remaining of 15\% of the images are for model interpretability in our proposed hybrid CNN-interpreter. 

\subsection{Model selection}
In this study, two alternative CNNs, namely, AlexNet and ResNet, are selected to demonstrate the detailed and consistent interpretabilities of our proposed methods. The model structures of AlexNet and ResNet are represented in Table \ref{tab:my-table} and \ref{tab:my-table2} respectively. 

\begin{table}[]
\caption{The network structure of AlexNet model}
\label{tab:my-table1}
\begin{tabular}{|l|l|l|}
\hline
Layer type          & Size of output feature map & Number of filters \\ \hline
Conv\_2d            & 59*59                      & 64                \\ \hline
Bn                  & 59*59                      & 64                \\ \hline
Max\_pooling\_2d    & 29*29                      & 64                \\ \hline
Conv\_2d\_1         & 29*29                      & 64                \\ \hline
Bn\_1               & 29*29                      & 64                \\ \hline
Max\_pooling\_2d\_1 & 14*14                      & 64                \\ \hline
Conv\_2d\_2         & 14*14                      & 64                \\ \hline
Bn\_2               & 14*14                      & 64                \\ \hline
Conv\_2d\_3         & 14*14                      & 64                \\ \hline
Bn\_3               & 14*14                      & 64                \\ \hline
Conv\_2d\_4         & 14*14                      & 64                \\ \hline
Bn\_4               & 14*14                      & 64                \\ \hline
Max\_pooling\_2d\_2 & 7*7                        & 64                \\ \hline
\end{tabular}
\end{table}


\begin{table}[]
\caption{The network structure of ResNet model}
\label{tab:my-table2}
\begin{tabular}{|l|l|l|l|}
\hline
Stage                    & Operations          & Size of output feature map & Number of filters \\ \hline
\multirow{4}{*}{Stage 1} & Conv\_1             & 120*120                    & 64                \\ \cline{2-4} 
                         & Bn\_1               & 120*120                    & 64                \\ \cline{2-4} 
                         & Activation          & 120*120                    & 64                \\ \cline{2-4} 
                         & Max\_pooling\_2d    & 59*59                      & 64                \\ \hline
\multirow{4}{*}{Stage 2} & Stage2\_input       & 59*59                      & 64                \\ \cline{2-4} 
                         & Convolutional block & 59*59                      & 256               \\ \cline{2-4} 
                         & Identity block 1    & 59*59                      & 256               \\ \cline{2-4} 
                         & Identity block 2    & 59*59                      & 256               \\ \hline
\multirow{5}{*}{Stage 3} & Stage3\_input       & 30*30                      & 128               \\ \cline{2-4} 
                         & Convolutional block & 30*30                      & 512               \\ \cline{2-4} 
                         & Identity block 1    & 30*30                      & 512               \\ \cline{2-4} 
                         & Identity block 2    & 30*30                      & 512               \\ \cline{2-4} 
                         & Identity block 3    & 30*30                      & 512               \\ \hline
\multirow{7}{*}{Stage 4} & Stage4\_input       & 15*15                      & 256               \\ \cline{2-4} 
                         & Convolutional block & 15*15                      & 1024              \\ \cline{2-4} 
                         & Identity block 1    & 15*15                      & 1024              \\ \cline{2-4} 
                         & Identity block 2    & 15*15                      & 1024              \\ \cline{2-4} 
                         & Identity block 3    & 15*15                      & 1024              \\ \cline{2-4} 
                         & Identity block 4    & 15*15                      & 1024              \\ \cline{2-4} 
                         & Identity block 5    & 15*15                      & 1024              \\ \hline
\multirow{4}{*}{Stage 5} & Stage5\_input       & 8*8                        & 512               \\ \cline{2-4} 
                         & Convolutional block & 8*8                        & 2048              \\ \cline{2-4} 
                         & Identity block 1    & 8*8                        & 2048              \\ \cline{2-4} 
                         & Identity block 2    & 8*8                        & 2048              \\ \hline
\end{tabular}
\end{table}

\subsection{Local interpretability for CNN-based Models}
The hybrid CNN-interpreter enables local interpretability with individual repeatable capacity by building a set of stacked ensemble submodels. Each submodel is developed by connecting the output of each layer in the original model and the final probability mapping layer to output prediction results, representing the model's learning ability. For the AlexNet model, the local interpretability will focus on each layer. For the ResNet model, the local interpretability will concentrate on each stage and component in each stage. The hybrid CNN interpreter will collect this set of prediction results for individual predictions. 

In our experiments, four different samples are input into the AlexNet model to explore the local interpretability, where we pick up both tumour and nontumour images as well as both the final correct prediction and wrong prediction as shown in the Table \ref{Alex_Result}.  

\begin{table}[ht]
\begin{tabular}{|l|l|l|l|}
\hline
Figures  & Tumor image   & Nontumor image & Result        \\ \hline
Figure \ref{local_alex} (a) & \faCheck      &                 & \faTimes      \\ \hline
Figure \ref{local_alex} (b) & \faCheck      &                 & \faCheck      \\ \hline
Figure \ref{local_alex} (c) &               & \faCheck        & \faTimes      \\ \hline
Figure \ref{local_alex} (d) &               & \faCheck        & \faCheck      \\ \hline
\end{tabular}
\label{Alex_Result}
\end{table}

\begin{figure*}
\centering
\includegraphics[width=1.1\textwidth]{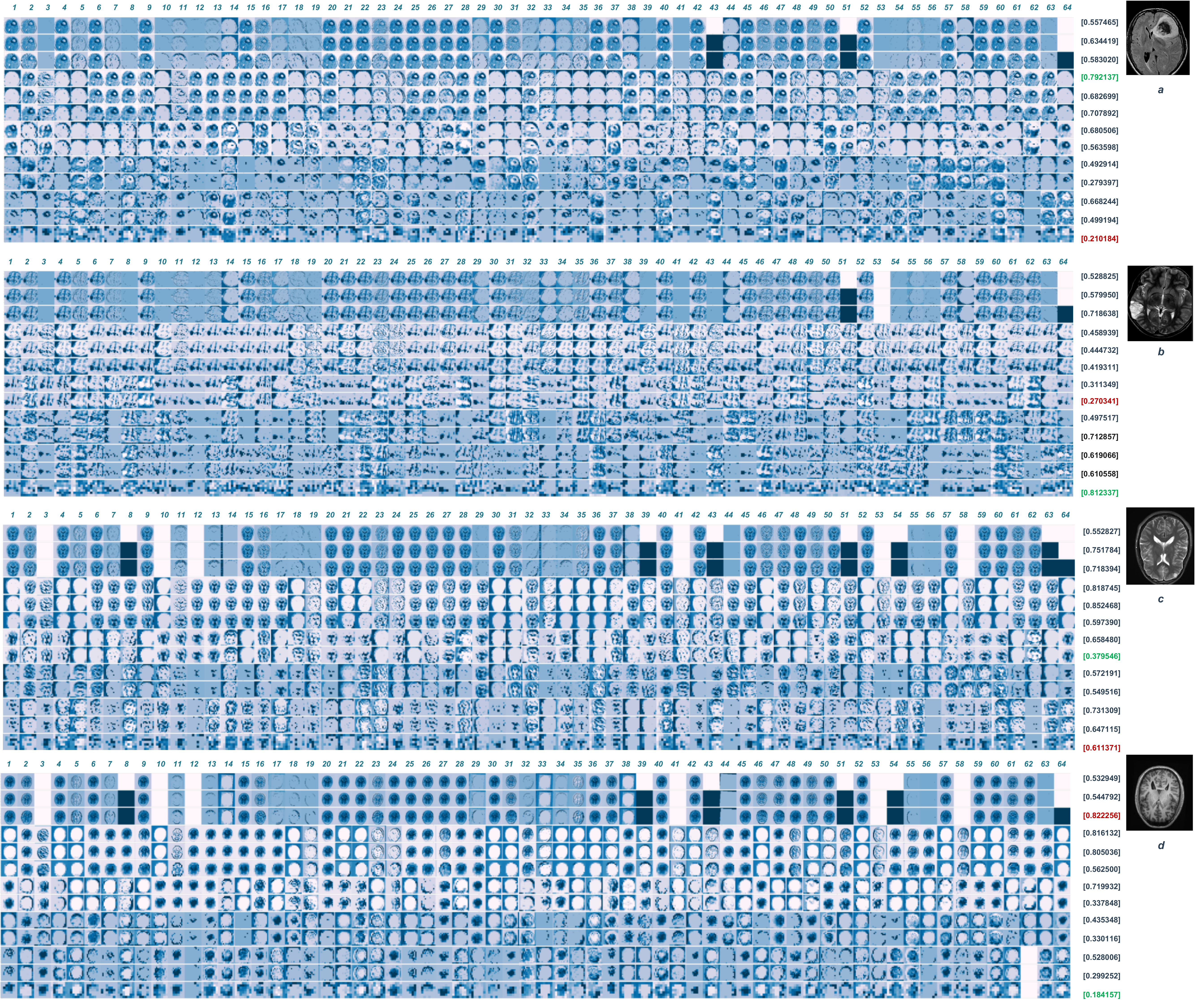}
\caption{\textbf{Local interpretability results of AlextNet model}. \textbf{a},Wrong prediction results of brain tumour.\textbf{b},Correct prediction results of brain tumour.\textbf{c},Wrong prediction results of non-brain tumour.\textbf{d},Correct prediction results of non-brain tumour}
\label{local_alex}
\end{figure*}

Figure \ref{local_alex} represents the prediction results of each layer in the AlexNet model, which is guaranteed consistent for repeatable running. From the local interpretability, we can see that each layer of different samples makes different individual predictions. Moreover, it is not an actual rule "more layers = better performance". For example, the $Conv2d\_1$ layer demonstrates the best performance in Figure (a) and the $Batch\_normalization\_2$ layer achieved the better performance in Figure 4(c).

\begin{figure*}[ht]
\centering
\includegraphics[width=1.1\textwidth]{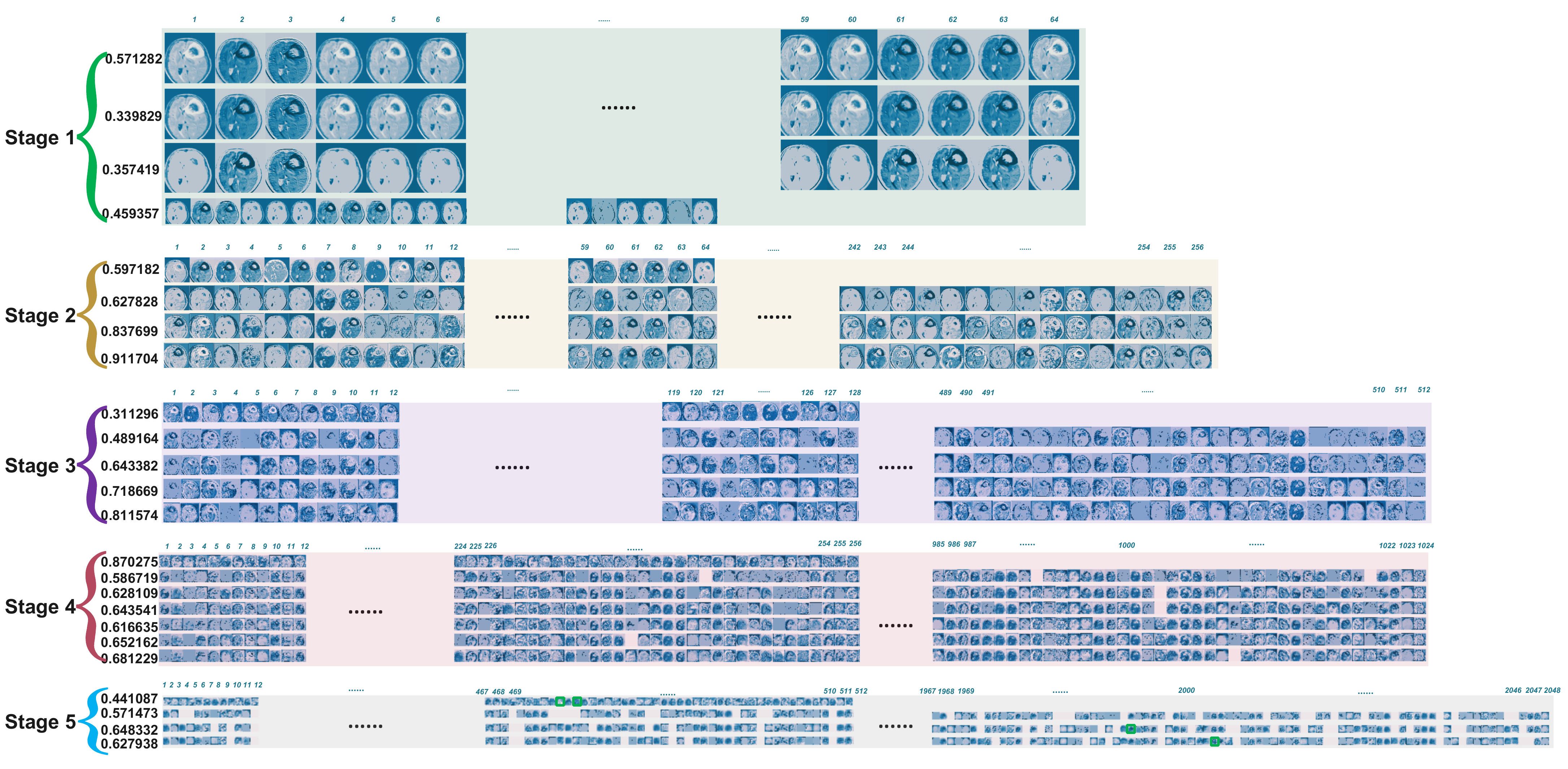}
\caption{\textbf{Local interpretability results of ResNet Model}: tumor image with final correct prediction results}
\label{tumor_2_ResNet}
\end{figure*}

Additionally, we interpret the ResNet model by computing a set of prediction results for each stage and each block as well. As shown in Figure \ref{tumor_2_ResNet}, the ResNet model shows that stages two and three have better performance compared with other stages in this test sample. For each internal stage, our proposed local interpretability can also provide a shortcut how learning by jumping over different residual blocks. For example, stage 5 is composed of one convolutional block and two identity blocks. The best performance in this stage was achieved by connecting the input layer and the first identity block.

\subsection{Global interpretability for CNN-based Models}

By combining the local interpretability of each layer across the entire training data, we present the regression-based methods to provide global insight into each filter in each layer, which will capture the global pattern of filters and the relationship between feature representations and prediction results. The experiments from the hybrid CNN-interpreter for global interpretability cover: 1) summary plots of the linear coefficient to represent the strength and direction of the linear relationship between each filter and prediction results in each layer, stage or internal block; 2) correlation analysis through layers of the AlexNet model and stages/blocks of the ResNet model respectively; 3) Filter importance analysis to reveal the details of positive and negative feature representations. 

\begin{figure*}[ht]
\centering
\includegraphics[width=1.0\textwidth]{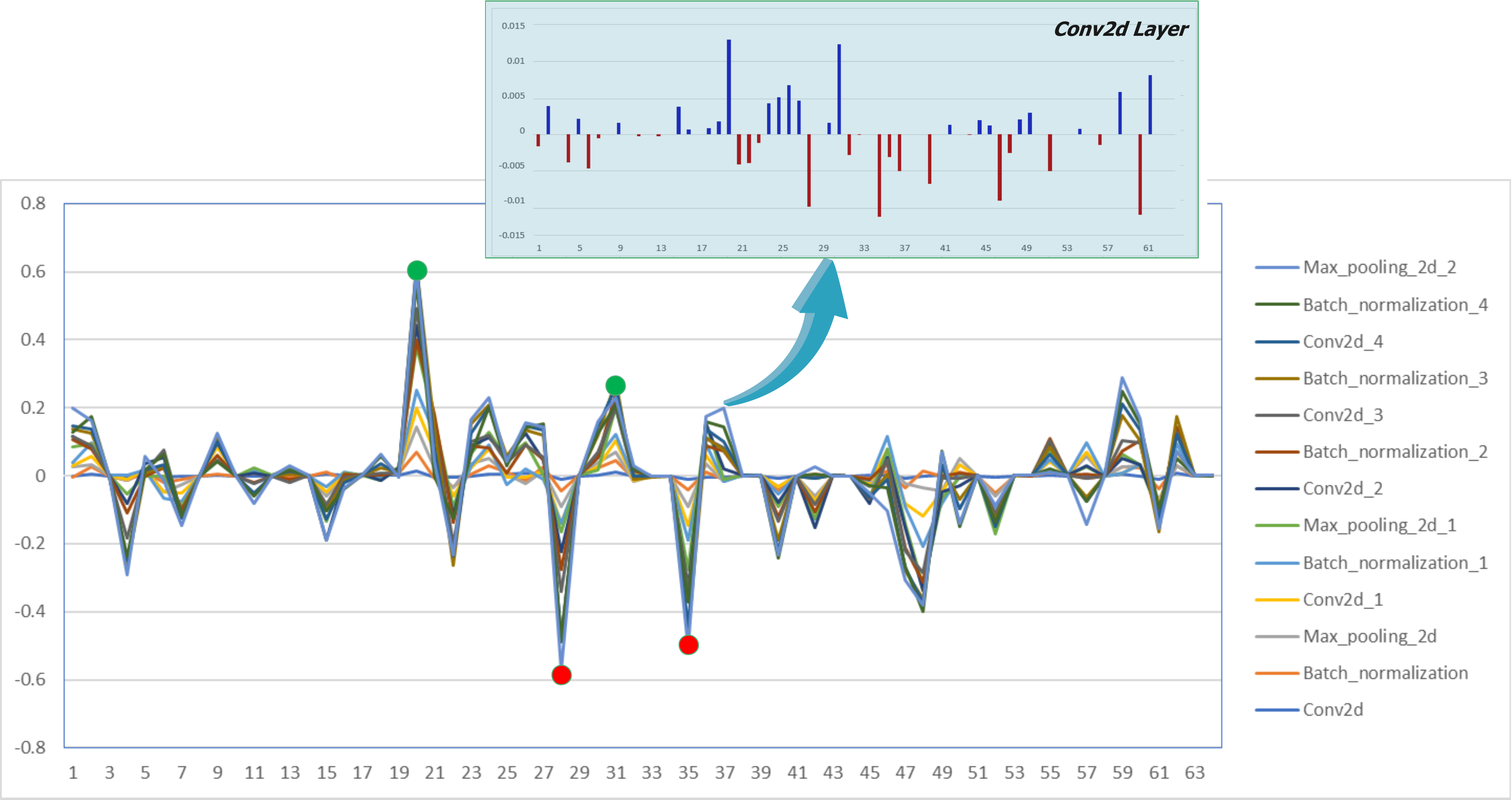}
\caption{\textbf{Summary plot of linear coefficients in AlextNet Model }}
\label{alexNet_regression}
\end{figure*}

We summarize the linear coefficients of all layers in the AlexNet model as shown in Figure \ref{alexNet_regression}, and the distribution of each filter’s coefficient in each layer can also be displayed by highlighting the positive as blue and the negative as red. Based on the results, we can see that filters 20 and 31 show the majority of positive effects among all the layers, whereas filters 28 and 35 show negative effects in most layers. 

We also get the linear coefficients of all stages and every internal block in the ResNet model.The overall distribution of coefficients among all stages can be shown in Figure \ref{resnet_regression} (a), and by using stage 4 as a reference, the positive and negative coefficients can be displayed in Figure \ref{resnet_regression} (b). Furthermore, to explore richer information, we can set a small range of filters such as filters 256-288 in stage 4, some consistent patterns among all the internal blocks can be revealed as shown in Figure \ref{resnet_regression} (c), filter 263 is positive in most blocks, and filters 259, 270, 284, and 285 usually are negative in majority internal blocks. This type of information can provide evidence for deriving lightweight models by eliminating layers, stages, or filters.

\begin{figure*}
\centering     
\includegraphics[width=1.0\textwidth]{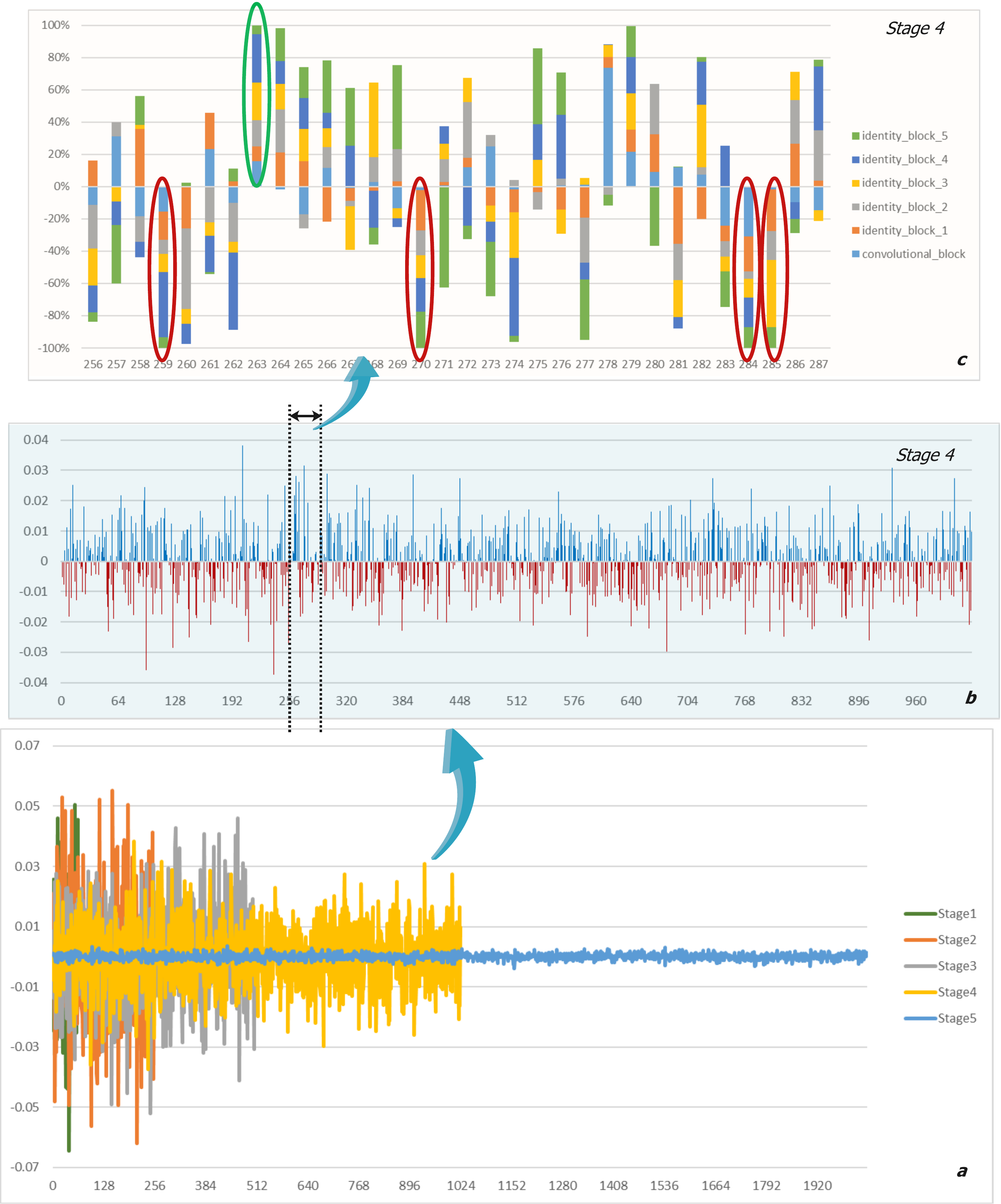}
\caption{\textbf{Summary plot of linear coefficients in ResNet Model}. \textbf{a},Linear coefficients of all stages. \textbf{b},Positive and negative coefficient distribution in stage 4. \textbf{c}, Coefficient patterns of internal blocks in stage 4 by setting a specific filter range.}
\label{resnet_regression}
\end{figure*}

\begin{figure*}
\centering     
\includegraphics[width=1.1\textwidth]{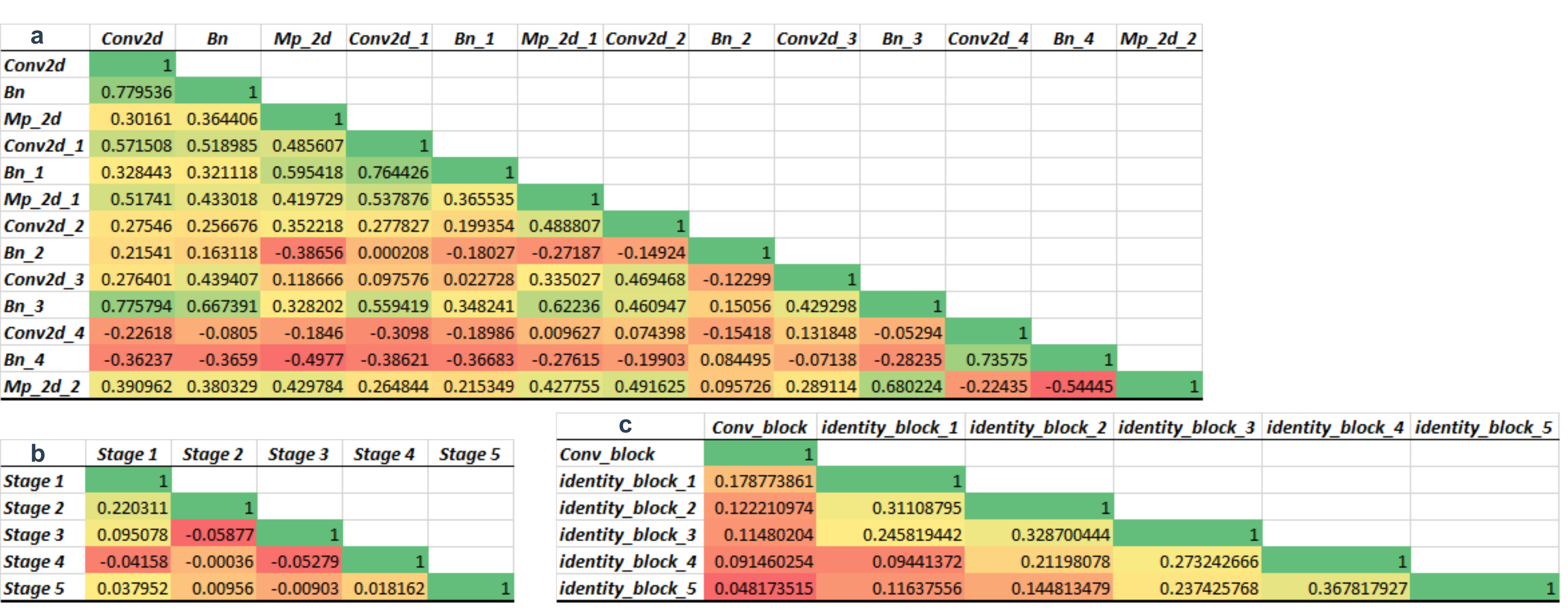}
\caption{\textbf{Correlation analysis among layers, stages or blocks for global interpretability}. \textbf{a},Correlation coefficients between layers in AlexNet Model. \textbf{b},Correlation coefficients between stages in ResNet Model.\textbf{c},Correlation coefficients between internal blocks in ResNet Model (by using stage 4 as reference).}
\label{correlation}
\end{figure*}

Based on the linear coefficient results above, we analyze the correlation between layers in the AlexNet model and the correlation between stages and internal blocks in each stage. Figure \ref{correlation} (a), we shows that the pairwise layers such as $Conv2d$ and $Batch\_normalization(Bn)$, $Conv2d\_1$ and $Batch\_normalization\_1(Bn\_1)$, etc. always have strong correlation with each other, which can reflect the local interpretability that these pairwise layers get the similar prediction results as well. Moreover, we can clearly see the $Con2d\_4$ and $Bn\_4$ layers have global negative effects for the final prediction results, compared with the local interpretability in Figure\ref{local_alex}, the prediction results in the $Max\_pooling\_2d\_2$ layer almost get noticeably different results compared with $Conv2d\_4$ layer and $Bn\_4$ layers, which can approve the consistency between local and global interpretability. 

For the ResNet model, Figure \ref{correlation} (b) shows that $stage 4$ has negative effects from $stage 1$ to $stage 3$. This reflects the consistency with the local interpretability in Figure \ref{tumor_2_ResNet} that the performance increased from $stage 1$ to $stage 3$ but dropped starting from the $stage 4$. Moreover, the correlations among the internal blocks in $stage 4$ indicate a generally positive correlation with each residual block. However, the correlations get less and less along with the blocks getting deeper, such as the correlation between $Conv_block$ and following identity blocks, is getting lower as shown in Figure \ref{correlation} (c).

\begin{figure*}[ht]
\centering
\includegraphics[width=0.8\textwidth]{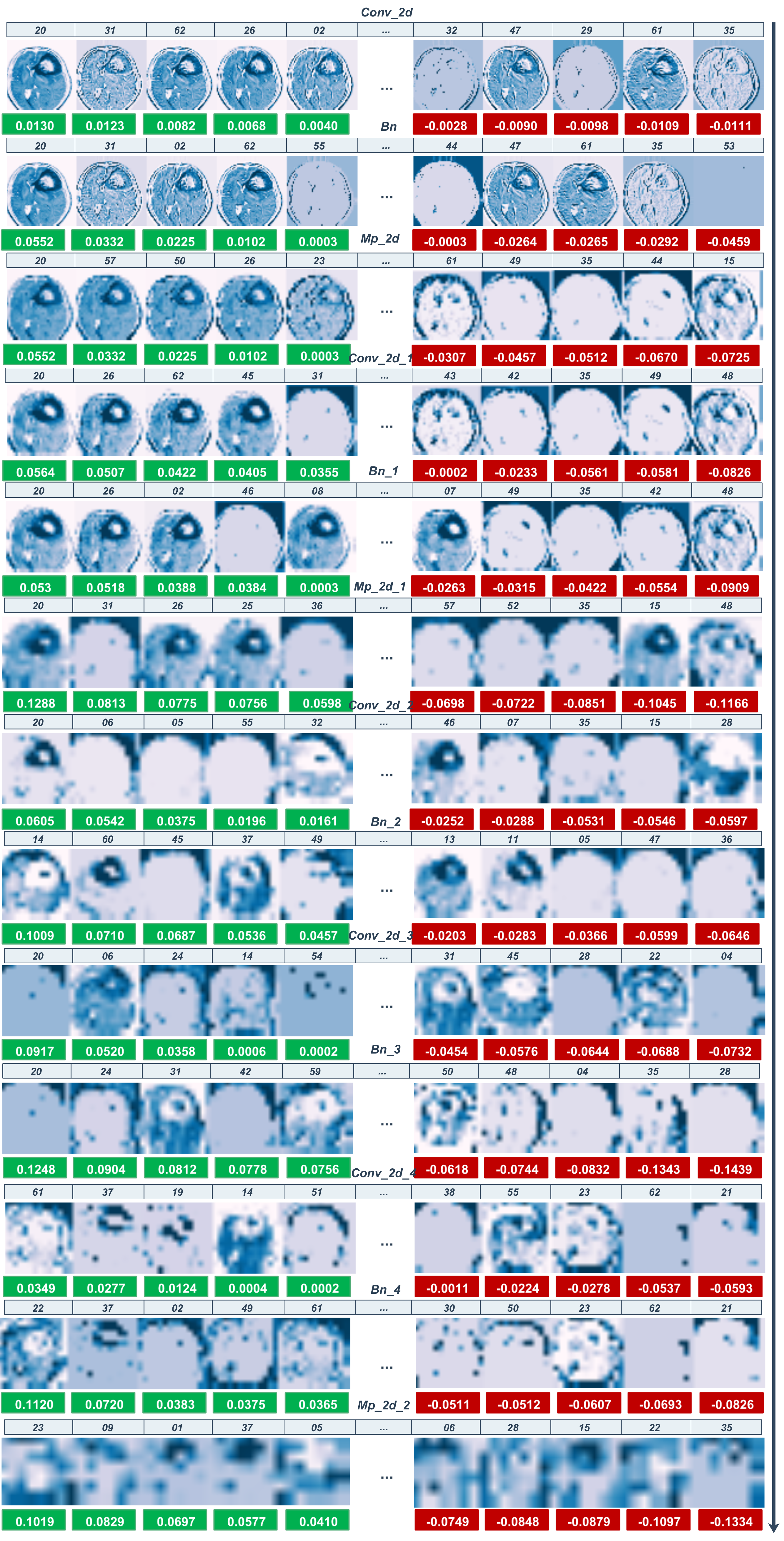}
\caption{Summary of filter importance analysis for AlexNet model}
\label{alexNet_top5_low5}
\end{figure*}

\begin{figure*}[ht]
\centering
\includegraphics[width=1.1\textwidth]{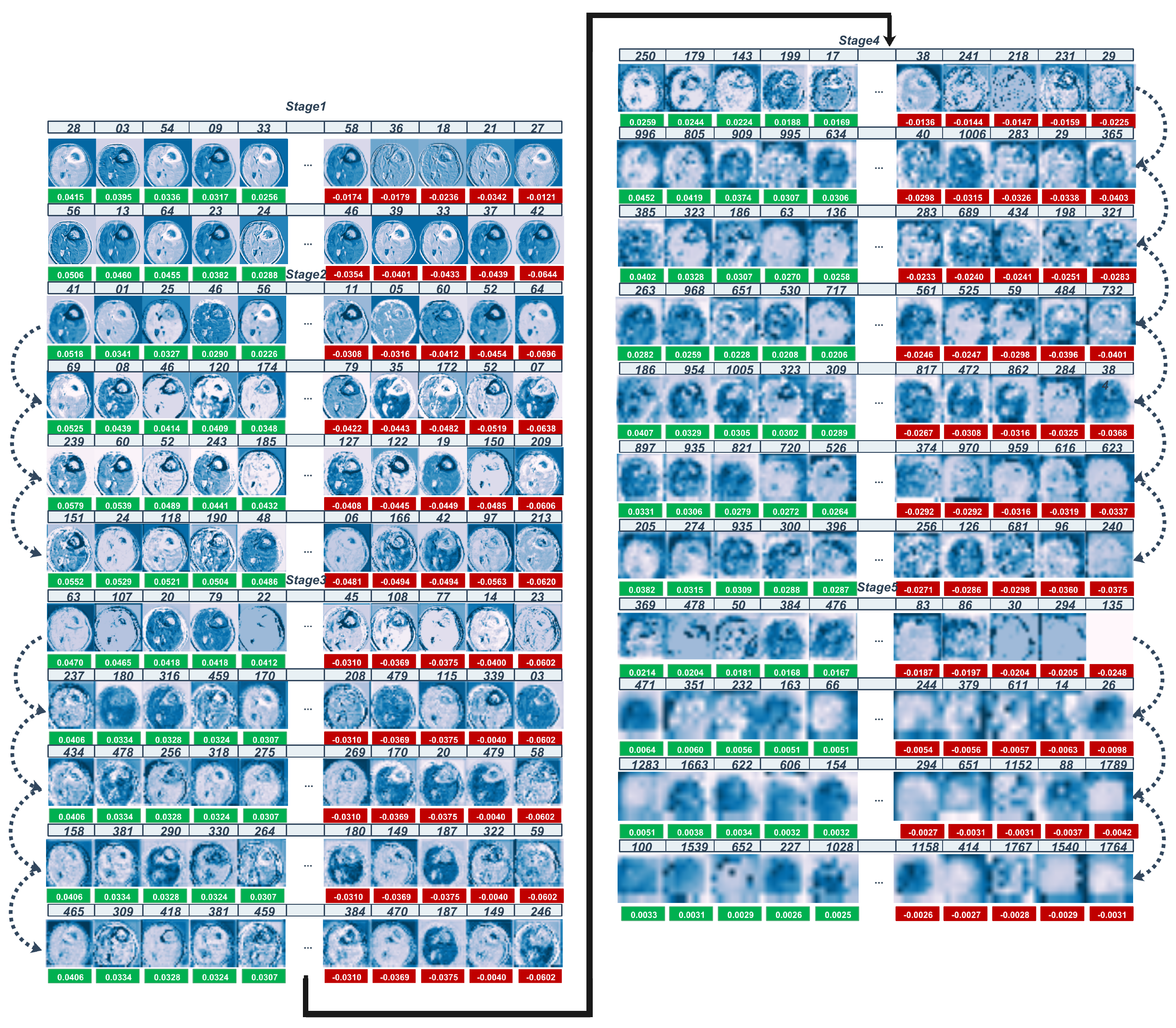}
\caption{Summary of filter importance analysis for ResNet model}
\label{resNet_top5_low5}
\end{figure*}

Finally, we extract the most essential filters in each layer to reveal the details of positive and negative feature representations. Figure \ref{alexNet_top5_low5} shows each layer's most significant five positives and five negatives using the same sample image in Figure \ref{local_alex} (a). From Figure \ref{alexNet_top5_low5}, we can see the different filters positively and negatively impact on different layers. For example, filter 20 positively impacts a total of nine layers that always focus on lower and middle layers, whereas filter 37 shows a positive impact on deeper layers. Filters 35 and 28 have a negative impact throughout the whole learning process. Additionally, the pairwise learning layers (pair of convolutional layer and batch normalization layer) show similar feature representations, which is consistent with our local interpretability. Figure \ref{resNet_top5_low5} summarises the filter importance analysis for the ResNet model. We pick up the input layer and output layer for each stage as well as the output layer for each internal block to show the feature map representations and their linear coefficient. Since the ResNet model is much more deeper than the AlexNet model, we can see that the deeper the layer, the more ambiguous the feature maps and the more difficult it is for the humans to understand their semantic information. 

By combining local interpretability and global interpretability, the experiments show that when the forward probability of the submodel is high, the features learned by the selected top five positive filters are consistent with human cognition with more significant semantic information, such as both the five filters (20,26,62,45,and 31) of $Conv\_2d\_1$ layer in AlexNet model and the filters (151,24,118,190,and 48) of $identity\_block\_2$ output layer in ResNet model showing the part of important features in the tumor area. Moreover, when the forward probability of the submodel is low, filters marked as negative effects can learn valuable semantic information. For example, the feature representation of filter 15 of the $Mp\_2d\_2$ layer in the AlexNet model actually covers part of the tumor area, and the filter 45 of the input layer in $stage 3$ in the ResNet model can represent parts of features in the tumor area as well. In contrast, filters considered positive by the model cannot learn the  critical parts of the image (e.g., 42 and 59 filters in $Bn\_3$ layer in the AlexNet model as well as 63 and 107 filters of the input layer in $stage 3$) can be understood that wrong cognition of the model leads to a low forward probability result. These could also verify the validity and consistency of our hybrid CNN-interpreter between the local and global interpretabilities. 

\section{Conclusion and Future Work}

The hybrid CNN interpreter helps users to have a deep conceptual understanding of CNN-based models by providing both local and global interpretabilites. The local interpretability by using original forward propagation can reveal how image data progresses through the layers of the CNN-based models, whereas the global filter importance based on the linear regression module can indicate how much each filter contributes to the model prediction.

The hybrid CNN interpreter can be widely used in different computer vision tasks, such as classification, object detection, and segmentation. The proposed interpreter can be flexibly adapted to various CNNs (layer-based, stage-based, or internal-block-based). The correlations between layers, stages, or blocks can also be generalized to understand the context of complex CNN structures. By demonstrating how to apply the hybrid CNN interpreter to explain different types of CNN-based models, we take brain tumor classification tasks to provide an interpretation of the AlexNet and ResNet models. The experiment results showed the efficiency and consistency between local and global interpretabilities, as well as among different models.

The proposed interpreter can be used to improve and debug models and to make CNN-based models more accurate and reliable. The local interpreter can reveal prediction results for specific samples of different learning processes, and based on these interpretabilities, we can set the gating and memory mechanisms to debug models that can help us identify where to access that memory and where to ignore it during the whole learning process. For example, the model can skip some layers with bad performances and tell where to access and make connections again. The feature correlations and filter importance identified by the global interpreter enable developers to determine which filter can be eliminated to get better performance.

\bibliographystyle{unsrt}  
\bibliography{references}

\end{document}